\documentclass[letter]{ieice}
\usepackage[pdftex]{graphicx}
\usepackage[fleqn]{amsmath}
\usepackage{amsthm}
\usepackage{newtxtext}
\usepackage[varg]{newtxmath}
\usepackage{bm}
\usepackage{mathtools}
\usepackage{cite}
\usepackage{here}

\field{A}
\title{Spectral Concentration at the Edge of Stability: Information Geometry of Kernel Associative Memory}
\authorlist{
  \authorentry{Akira Tamamori}{m}{AIT}\MembershipNumber{1512145}
  \affiliate[AIT]{The author is with the Department of
    Information Science, Aichi Institute of Technology, Toyota-shi, 470-0392 Japan.}
}
\received{2015}{1}{1}
\revised{2015}{1}{1}

\begin{document}
\maketitle

\begin{summary}
  High-capacity kernel Hopfield networks exhibit a \textit{Ridge of
    Optimization} characterized by extreme stability. While previously
  linked to \textit{Spectral Concentration}, its origin remains
  elusive. Here, we analyze the network dynamics on a statistical
  manifold, revealing that the Ridge corresponds to the Edge of
  Stability, a critical boundary where the Fisher Information Matrix
  becomes singular. We demonstrate that the apparent Euclidean force
  antagonism is a manifestation of \textit{Dual Equilibrium} in the
  Riemannian space. This unifies learning dynamics and capacity via
  the Minimum Description Length principle, offering a geometric
  theory of self-organized criticality.
\end{summary}

\section{Introduction}
Associative memory is fundamentally a geometric problem: how to embed
discrete patterns as stable fixed points within a continuous state
space. Recent advances using Kernel Logistic Regression (KLR) have
demonstrated that learning can sculpt these landscapes to achieve
capacities far exceeding classical limits~\cite{tamamori2025,
  tamamori2025b, Gardner1988}.  Our previous phenomenological analysis
identified a \textit{Ridge of Optimization} where stability is
maximized via a mechanism we termed \textit{Spectral Concentration},
defined as a state where the weight spectrum exhibits a sharp
hierarchy~\cite{tamamori2025c}.

However, a deeper question remains: \textit{Why} does the learning
dynamics self-organize into this specific spectral state? Why does the
system operate at the brink of instability?  To answer these
questions, we must look beyond the Euclidean geometry of the weight
parameters and consider the intrinsic geometry of the probability
distributions they represent. This is the domain of Information
Geometry~\cite{Amari2016}.

In this work, we reinterpret the KLR Hopfield network as a statistical
manifold equipped with a Fisher-Rao metric. We show that:
\begin{enumerate}
\item The Ridge is the locus where the Fisher Information Matrix (FIM)
  undergoes a spectral phase transition, effectively minimizing the
  information dimension of the memory.
\item The optimization dynamics on the Ridge satisfies a \textit{Dual
    Equilibrium}, where the massive Euclidean gradients are
  neutralized by the curvature of the manifold, resulting in a
  stationary state in the dual affine coordinates.
\end{enumerate}

Although our focus is on associative memory, the geometric mechanisms
revealed here, specifically Spectral Concentration and Dual
Equilibrium, may offer broader insights into the generalization
capabilities of over-parameterized neural networks, where similar
phenomena of ``feature learning'' and ``sharpness selection'' have
been reported. Rather than introducing an entirely new learning
principle, this work reinterprets the Ridge of Optimization as an
emergent boundary where existing geometric quantities, such as the
Fisher Information Matrix and natural gradient dynamics, become
maximally expressed.

\section{Geometric Framework}
\label{sec:framework}

We consider the kernel Hopfield network not merely as a dynamical
system, but as a parametric statistical model.

\subsection{Statistical Manifold of Kernel Memory}
Let $\mathcal{H}$ be a reproducing kernel Hilbert space with kernel
$K(\cdot, \cdot)$. The state of the network is determined by the dual
variables $\boldsymbol{\alpha} \in \mathbb{R}^{P \times N}$, where $P$
is the number of stored patterns and $N$ is the number of neurons. For
a single neuron $i$, the probability of being in state $s_i=+1$ given
the input pattern $\boldsymbol{\xi}^\mu$ is modeled by the logistic
sigmoid function $\sigma(\cdot)$:
\begin{equation}
    p(\boldsymbol{\xi}^\mu; \boldsymbol{\alpha}_i) = \sigma \left( \sum_{\nu=1}^P \alpha_{\nu i} K(\boldsymbol{\xi}^\mu, \boldsymbol{\xi}^\nu) \right).
\end{equation}
The set of all such realizable probability distributions forms a
statistical manifold
$\mathcal{M} = \{ p(\cdot; \boldsymbol{\alpha}) \mid
\boldsymbol{\alpha} \in \mathbb{R}^{P \times N} \}$.

\subsection{Fisher Information Matrix (FIM)}
The intrinsic geometry of $\mathcal{M}$ is governed by the Fisher
Information Matrix (FIM) $G(\boldsymbol{\alpha})$. For the weights
associated with a single neuron (omitting index $i$), the FIM is given
by:
\begin{equation}
    G_{\mu\nu}(\boldsymbol{\alpha}) = \mathbb{E} \left[ \frac{\partial \log p}{\partial \alpha_\mu} \frac{\partial \log p}{\partial \alpha_\nu} \right].
\end{equation}
In the context of KLR, this takes the specific form involving the kernel Gram matrix $\bm{K}$:
\begin{equation}
    G(\boldsymbol{\alpha}) = \bm{K} \bm{D}(\boldsymbol{\alpha}) \bm{K},
    \label{eq:fim_kernel}
\end{equation}
where $\bm{D}(\boldsymbol{\alpha})$ is a diagonal matrix with
entries~$D_{\mu\mu} = p_\mu (1 - p_\mu)$ representing the variance of
the prediction for pattern $\mu$.

\subsection{Riemannian vs. Euclidean Gradients}
The learning dynamics can be viewed as a flow on this manifold. The
standard gradient descent follows the steepest direction in the
Euclidean metric,
$\Delta \boldsymbol{\alpha} \propto -\nabla_{\boldsymbol{\alpha}}
L$. However, the intrinsic steepest direction is given by the Natural
Gradient~\cite{Amari1998}, which accounts for the curvature $G$:
\begin{equation}
    \tilde{\nabla} L = G^{-1} \nabla_{\boldsymbol{\alpha}} L.
\end{equation}
Our central hypothesis is that the Ridge phenomena emerge from the
interplay between these two vector fields---the Euclidean force
$\nabla L$ and the geometric curvature $G$.

\section{Spectral Concentration as Information Compression}
\label{sec:spectral_concentration}

Our previous phenomenological study identified \textit{Spectral
  Concentration} in the weight matrix $\boldsymbol{\alpha}$ as the key
to stability~\cite{tamamori2025c}. Here, we show that this is a direct
consequence of the information geometry of the learned manifold.

\subsection{Effective Dimensionality of Memory}
\begin{figure}[t]
\begin{center}
  \includegraphics[width=\hsize]{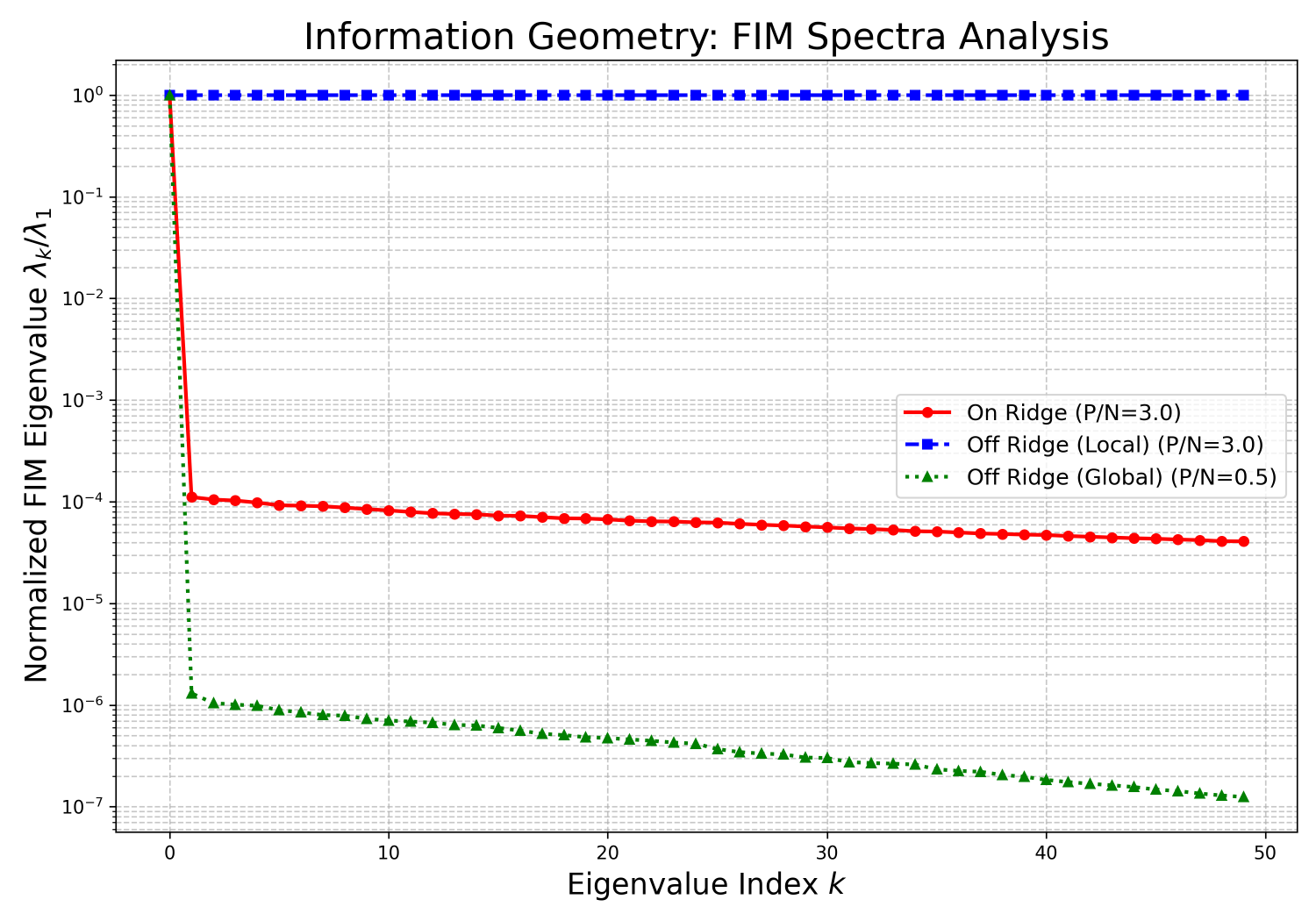}
  \caption{Spectrum of the Fisher Information Matrix $G$ (normalized
    by $\lambda_1$).  \textbf{Blue (Off Ridge, Local):} The spectrum
    is flat ($\lambda_k/\lambda_1 \approx 1$), indicating a lack of
    structural compression.  \textbf{Red (On Ridge):} The spectrum
    exhibits \textbf{Spectral Concentration}, dropping sharply after
    the first mode but maintaining a non-zero tail. This ``L-shaped''
    profile signifies optimal information compression.  \textbf{Green
      (Off Ridge, Global):} The tail collapses to near zero,
    indicating rank deficiency.}
  \label{fig:fim_spectrum}
\end{center}
\end{figure}

The rank of the Fisher Information Matrix $G$ determines the number of
independent directions in the probability space that the model can
represent locally. We analyze the eigenvalue spectrum of $G$, denoted
by $\lambda_1 \ge \lambda_2 \ge \dots \ge \lambda_P$.  We define the
\textit{Effective Information Dimension} $d_{\text{eff}}$ using the
stable rank:
\begin{equation}
  d_{\text{eff}}(G) = \dfrac{\left(\sum_{k=1}^{P} \lambda_k\right)^2}{\sum_{k=1}^{P} \lambda_k^2}.
\end{equation}
\textbf{Observation 1 (L-Shaped Spectrum):} Numerical analysis
(Fig.~\ref{fig:fim_spectrum}) reveals a striking contrast in the
spectral structures.  In the local regime (blue dashed line), the
spectrum is flat ($\lambda_k \approx \lambda_1$), indicating a diffuse
distribution of information where no dominant direction exists.  In
contrast, on the Ridge (red solid line), the spectrum exhibits a
characteristic ``L-shape'': the normalized eigenvalues drop by orders
of magnitude immediately after the leading mode
($\lambda_2 \ll \lambda_1$), yet the tail remains non-zero and flat
($\lambda_{k>1} > 0$).  This confirms that the Ridge achieves
stability via Spectral Concentration, effectively compressing noise
into a single dominant mode while preserving the dimensionality
required for memory capacity.  Quantitatively, while the local regime
exhibits $d_{\text{eff}}(G) \approx P$ (full utilization of
dimensions), the Ridge collapses this to
$d_{\text{eff}}(G)\approx 1.5 \sim 5.0$, depending on the load. This
drastic reduction confirms that the network performs optimal
compression, retaining only the essential degrees of freedom.

\subsection{Minimum Description Length (MDL) Principle}
This spectral structure can be interpreted through the Minimum
Description Length (MDL) principle~\cite{Rissanen1978}. The network
compresses the high-dimensional input noise into a single dominant
mode ($\lambda_1$) to maximize robustness, while maintaining just
enough degrees of freedom ($\lambda_{k>1}$) to separate the $P$ stored
patterns.  Unlike a Rank-1 collapse ($G \to \bm{v}_1\bm{v}_1^\top$),
where pattern distinguishability is lost ($\lambda_{k>1} \to 0$), the
Ridge maintains a ``working margin'' of information dimensionality.

\section{The Edge of Stability}
\label{sec:edge_of_stability}

Why does the learning dynamics stop at this state of Spectral
Concentration? We propose that the Ridge corresponds to the limit of
stability for the optimization dynamics.

\subsection{Maximal Curvature at the Ridge}
The local curvature of the statistical manifold is quantified by the
largest eigenvalue of the FIM, $\lambda_{\max}(G)$. In gradient
descent dynamics, the stability of the update steps is constrained by
this curvature; excessively sharp curvature can destabilize the
learning process, a phenomenon known as the \textit{Edge of
  Stability}~\cite{Cohen2021}.  We refer to the Ridge as an ``Edge of
Stability'' in this context, not as a new dynamical phase, but as a
geometric boundary characterized by the maximal sensitivity encoded in
the Fisher Information Matrix.

\begin{figure}[t]
\begin{center}
  \includegraphics[width=\hsize]{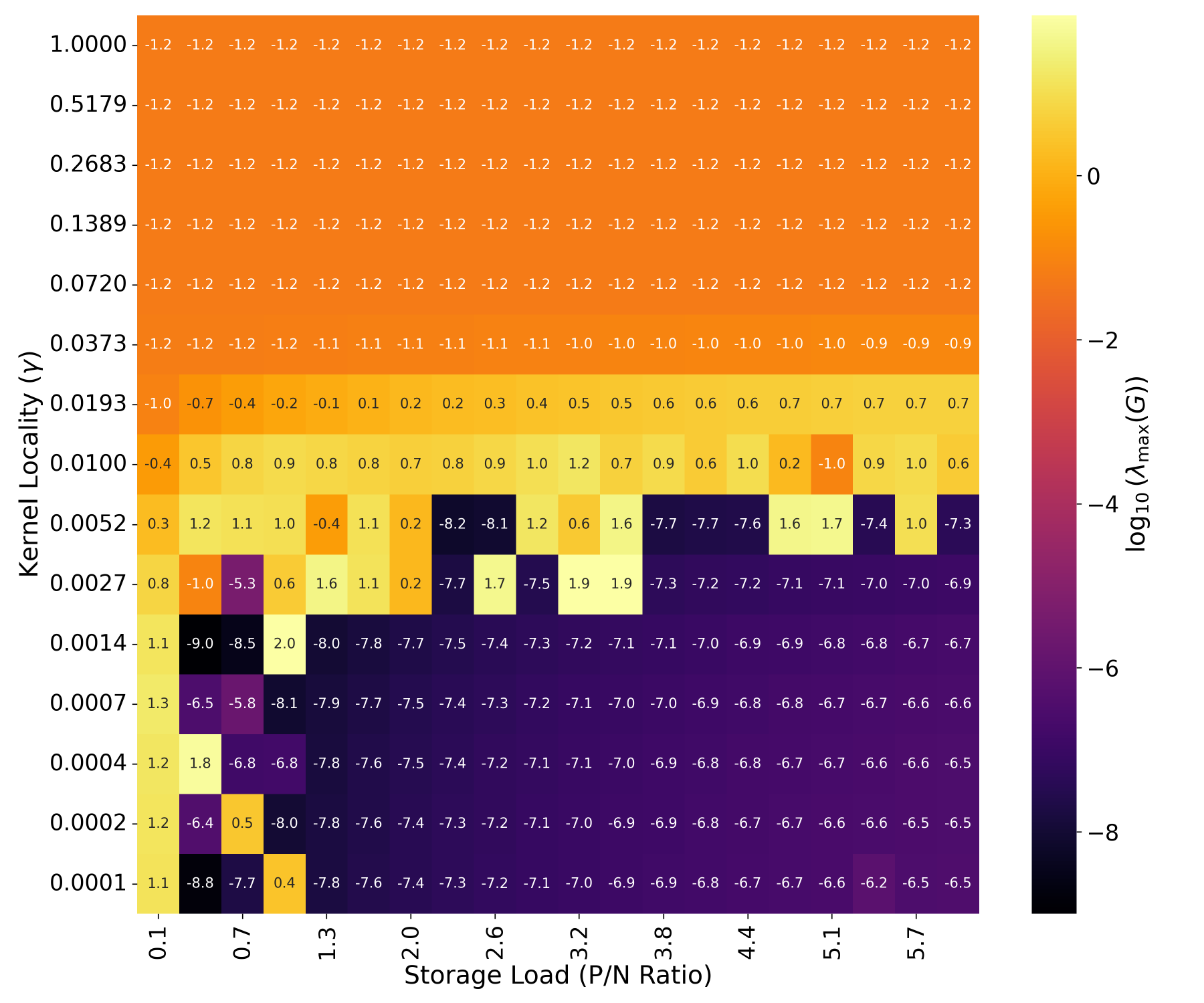}
  \caption{Phase diagram of the maximum eigenvalue of the Fisher
    Information Matrix, $\log_{10}(\lambda_{\max}(G))$ (averaged over 5 trials).  The Ridge
    (bright diagonal band) corresponds to the region where the
    manifold curvature is locally maximized ($\lambda_{\max} \gg 1$),
    effectively creating the ``Edge of Stability.''  In contrast, the
    dark region (bottom right) indicates information collapse due to
    saturation of the sigmoid function ($p \to 0/1$), leading to
    vanishing curvature.}
  \label{fig:eos_diagram}
\end{center}
\end{figure}

\textbf{Observation 2 (Criticality):} Our phase diagram analysis
(Fig.~\ref{fig:eos_diagram}) reveals that $\lambda_{\max}(G)$ is
maximized exactly on the Ridge.  While in the local regime (top
region), the curvature is relatively small
($\log_{10} \lambda_{\max} \approx -1.2$), on the Ridge, it reaches
significantly higher values ($\log_{10} \lambda_{\max} > 1.0$),
indicating the formation of extremely sharp attractors.  Crucially,
immediately beyond the Ridge (in the global regime, bottom right), the
value drops precipitously ($\log_{10} \lambda_{\max} < -6.0$). This
corresponds to a phase where the learning saturates ($p_\mu \to 0$ or
$1$), causing the Fisher information to vanish due to the vanishing
gradients of the sigmoid function.  Thus, the Ridge represents the
critical boundary where the system maximizes geometric curvature to
the limit allowed before information collapse occurs.

\subsection{The Cliff of Optimization}
Thus, the Ridge of Optimization can be geometrically defined as the
\textbf{Edge of Stability}: the manifold boundary where the curvature
is maximal.
\begin{equation*}
\text{Ridge} \approx \left\{ (\gamma, P/N) \mid \lambda_{\max}(G) \text{ is locally maximized} \right\}.
\end{equation*}
The network self-organizes to this cliff edge because sharper
attractors (higher curvature) provide stronger error correction. The
system climbs the curvature gradient until it reaches the limit
imposed by the learning dynamics or the saturation of the
nonlinearity.

\section{Dual Geometry and Force Antagonism}
\label{sec:dual_geometry}

Our previous work identified a phenomenological \textit{Force
  Antagonism} on the Ridge: the Direct Force $\bm{F}_d$ and Indirect
Force $\bm{F}_i$ become massive and strongly anti-correlated
($\bm{F}_d \approx -\bm{F}_i$). We now reveal the geometric origin of
this antagonism.

\subsection{Euclidean vs. Riemannian Forces}
The learning dynamics can be analyzed in two distinct metric spaces:
\begin{enumerate}
\item \textbf{Euclidean Space ($\mathbb{R}^P$):} The standard
  parameter space where weights $\boldsymbol{\alpha}$ reside. The
  driving force is the Euclidean gradient $\nabla L$.
    \item \textbf{Riemannian Space ($\mathcal{M}$):} The statistical
      manifold equipped with the Fisher metric $G$. The intrinsic
      driving force is the Natural Gradient
      $\tilde{\nabla} L = G^{-1} \nabla L$.
\end{enumerate}

\begin{figure}[t]
  \centering
  \begin{minipage}{\linewidth}
    \centering
    \includegraphics[width=\linewidth]{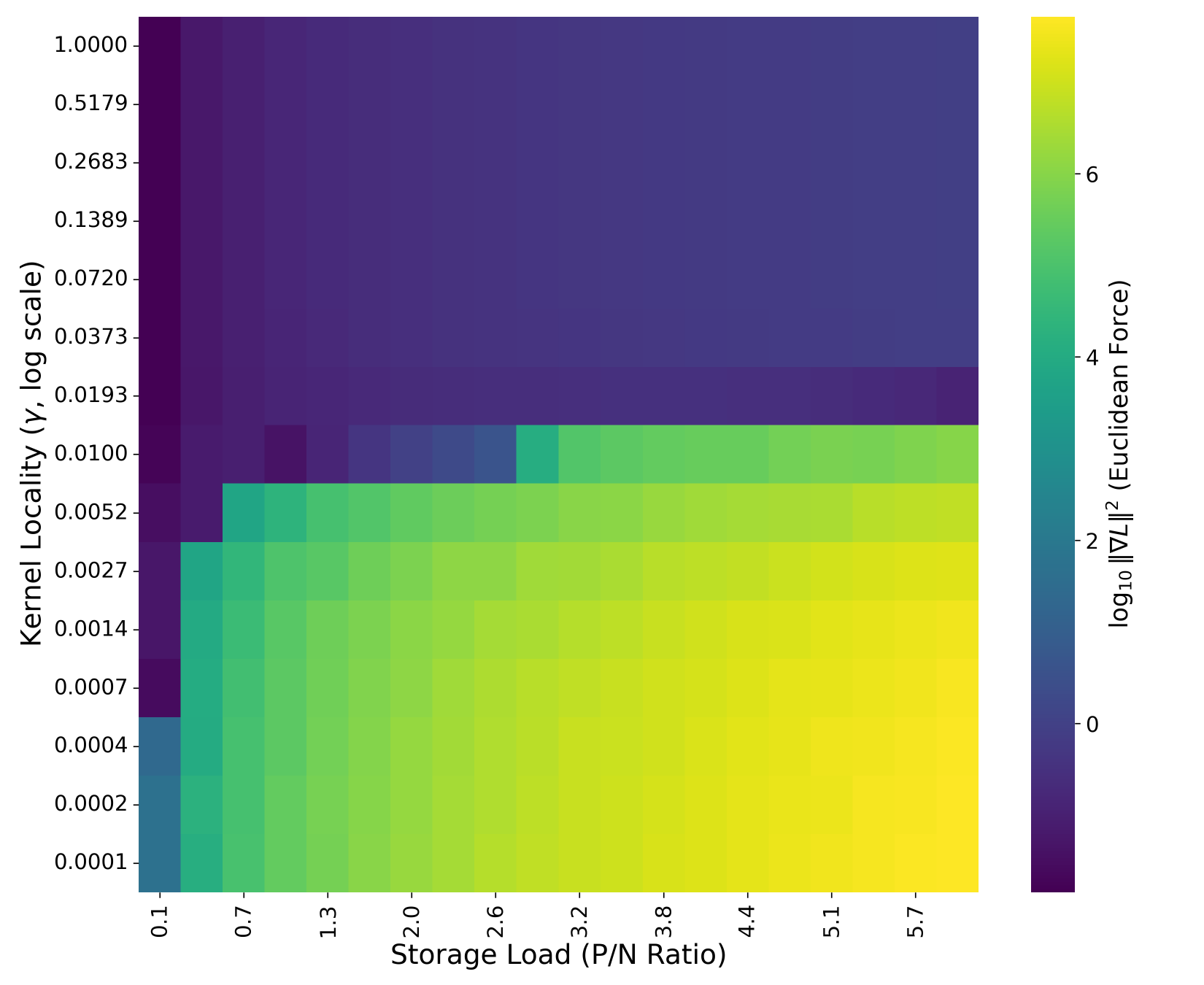}
    \text{(a): Euclidean Gradient Norm}
    \label{fig:dual_euclidean}
  \end{minipage}
  \hfill
  \begin{minipage}{\linewidth}
    \centering
    \includegraphics[width=\linewidth]{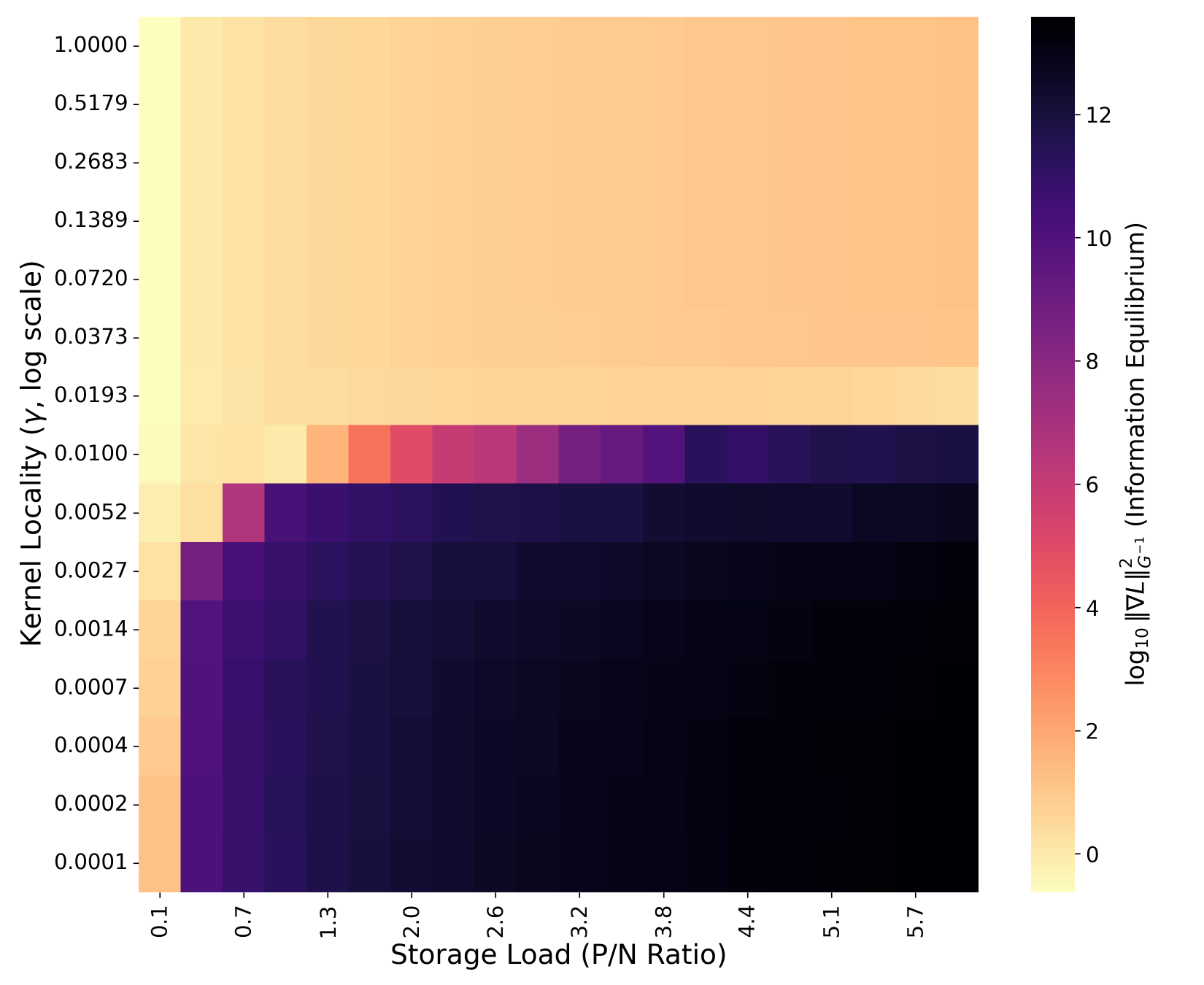}
    \text{(b): Riemannian Gradient Norm}
    \label{fig:dual_riemannian}
  \end{minipage}
  \caption{Comparison of Gradient Norms: (a) Euclidean Gradient
    $\|\nabla L\|^2$ vs. (b) Riemannian (Natural) Gradient
    $\|\tilde{\nabla} L\|_{G^{-1}}^2$ (averaged over 5 trials).  On
    the Ridge, the Euclidean norm diverges (bright band in a),
    reflecting the massive ``Force Antagonism.'' However, the
    Riemannian norm is minimized (dark band in b), revealing that the
    system is actually in a state of \textbf{Dual Equilibrium} with
    respect to the intrinsic geometry. This confirms that the apparent
    instability in parameter space is resolved by the curvature of the
    statistical manifold.}
  \label{fig:dual_balance}
\end{figure}

\textbf{Observation 3 (Dual Equilibrium):} Numerical experiments
(Fig.~\ref{fig:dual_balance}) demonstrate a striking contrast on the
Ridge:
\begin{itemize}
\item The Euclidean gradient norm $\|\nabla L\|^2$ diverges (bright
  region), indicating massive internal stresses.
    \item The Riemannian gradient norm
      $\|\nabla L\|_{G^{-1}}^2 = \nabla L^\top G^{-1} \nabla L$ is
      minimized (dark region), indicating that the system is close to
      an intrinsic equilibrium.
\end{itemize}

\subsection{Geometric Derivation of Antagonism}
We define the Indirect Force in the language of information geometry
as the correction term arising from the curvature of the manifold.
The Natural Gradient relation $\nabla L = G \tilde{\nabla} L$ implies
that the observed Euclidean force is the intrinsic force distorted by
the metric $G$.  On the Ridge, we found that the spectrum of $G$ is
dominated by a leading eigenvalue $\lambda_1 \gg 1$ with eigenvector
$\bm{v}_1$.  Approximating
$G \approx \lambda_1 \bm{v}_1 \bm{v}_1^\top$, the relationship
becomes:
\begin{equation}
    \nabla L \approx \lambda_1 (\bm{v}_1^\top \tilde{\nabla} L) \bm{v}_1.
\end{equation}
This explains the ``amplification'' mechanism: the huge curvature
$\lambda_1$ magnifies small intrinsic mismatches into massive
Euclidean gradients.

The decomposition $\nabla L = \bm{F}_d + \bm{F}_i$ can be understood
as splitting the gradient into the ``signal'' component and the
``curvature correction'' component. The condition
$\nabla L \to \bm{0}$ (equilibrium) in Euclidean space implies
$\bm{F}_d \approx -\bm{F}_i$.  However, the condition
$\tilde{\nabla} L \to \bm{0}$ in Riemannian space is much stronger and
more fundamental. The Ridge represents a state where the system has
found a \textbf{Dual Equilibrium}: a point where the intrinsic error
is minimized despite the extreme curvature of the ambient space.  This
dual equilibrium does not introduce a new equilibrium concept, but
highlights the coexistence of instability in the Euclidean parameter
space and stability in the information-geometric (Riemannian) sense.

Interestingly, the massive curvature $\lambda_1$ on the Ridge implies
that the Natural Gradient step in the principal direction, scaled by
$\lambda_1^{-1}$, becomes negligible. This suggests a self-stabilizing
mechanism: as the system approaches the Edge of Stability, the
intrinsic update step vanishes, naturally braking the learning
dynamics exactly at the optimal boundary.

\section{Discussion}
\label{sec:discussion}
In this study, we have reinterpreted the dynamics of KLR-trained
Hopfield networks through the lens of Information Geometry. Our
analysis reveals that the phenomenological Ridge of Optimization
is not merely a sweet spot of hyperparameters, but a manifestation of
fundamental geometric principles: \textit{Spectral Concentration},
\textit{Edge of Stability}, and \textit{Dual Equilibrium}.

\subsection{Self-Organized Criticality in Memory}
Our finding that the optimal memory state lies at the brink of
information collapse (Fig.~\ref{fig:eos_diagram}) is reminiscent of
the hypothesis of Self-Organized Criticality (SOC)~\cite{Beggs2003} in
biological neural networks, in the sense that the system naturally
tunes itself to a critical boundary. The brain is thought to operate
at a critical point to maximize information transmission and dynamic
range. Similarly, our KLR Hopfield network tunes itself to the Edge of
Stability, maximizing the curvature of the statistical manifold to
create deep attractors, while stopping just short of the singular
region where memory capacity vanishes. This suggests that SOC may be a
universal requirement for high-capacity associative memory.

\subsection{Implications for Deep Learning Generalization}
The phenomenon of Spectral Concentration may offer broader insights
into the generalization mechanics of over-parameterized neural
networks.  Recent studies report a ``feature learning'' regime, or
Neural Collapse~\cite{Papyan2020}, where the weight spectrum becomes
low-rank to capture task-relevant structure.  Notably, Karakida et
al.~\cite{Karakida2020, Karakida2021} demonstrated that deep neural
networks inherently exhibit a pathological FIM spectrum dominated by
outliers. Our findings suggest that KLR learning exploits this
spectral pathology to stabilize memory.
Our results provide a geometric explanation for this: the network
minimizes the effective dimensionality (via the Fisher Information
spectrum) to ensure robust inference (Dual Equilibrium), effectively
performing a geometric Occam's razor.

\subsection{The Geometry of Attention Mechanisms}
Modern Hopfield Networks are mathematically equivalent to the
attention mechanism in Transformers~\cite{Ramsauer2021}.  Our
framework suggests that attention layers may also rely on specific
spectral structures in their Key-Query matrices. If attention heads
operate at an Edge of Stability, our duality theory could provide new
tools for analyzing large-scale Transformers. Future work will
investigate whether they exhibit spectral signatures similar to the
Ridge.

\section{Conclusion}
This geometric framework unifies the apparently contradictory
observations of instability (high curvature, Edge of Stability) and
stability (deep attractors, Dual Equilibrium). The Ridge of Optimization
is not merely a parameter sweet spot, but a manifestation of
\textit{Self-Organized Criticality} in the information geometry of
neural networks. The network tunes itself to the brink of instability,
maximizing curvature to carve the deepest possible attractor basins,
while maintaining dual equilibrium to ensure learning convergence.

\end{document}